\documentclass[10pt,twocolumn,letterpaper]{article}

\usepackage[pagenumbers]{cvpr}

\usepackage{caption}

\usepackage[dvipsnames]{xcolor}

\usepackage{xspace}
\usepackage{comment}

\usepackage{amsmath}
\usepackage[frozencache=true,cachedir=.]{minted}
\usepackage{multirow}
\newcommand*{\affaddr}[1]{#1} 
\newcommand*{\affmark}[1][*]{\textsuperscript{#1}}
\newcommand*{\email}[1]{\small{\texttt{#1}}}

\definecolor{cvprblue}{rgb}{0.21,0.49,0.74}
\usepackage[pagebackref,breaklinks,colorlinks,citecolor=cvprblue]{hyperref}

\title{Reconstructing Animals {\em and} the Wild}

\author{
Peter~Kulits\affmark[1] \quad
Michael~J.~Black\affmark[1] \quad
Silvia~Zuffi\affmark[2]\\
{\small \affaddr{\affmark[1]Max Planck Institute for Intelligent Systems, T{\"u}bingen, Germany} \quad \affaddr{\affmark[2]IMATI-CNR, Milan, Italy}}\\
\tt\small{\{kulits,black\}@tue.mpg.de,\quad \email{silvia@mi.imati.cnr.it}
\vspace{-2.0em}
}
}

\begin{document}
\newcommand{\teaserCaption}{
We train an LLM to decode a frozen CLIP embedding of a natural image into a structured compositional scene representation encompassing both animals and their habitats.
}

\twocolumn[{
    \renewcommand\twocolumn[1][]{#1}
    \maketitle
    \centering
    \begin{minipage}{1.00\textwidth}
        \centering
        \includegraphics[width=0.16025\linewidth]{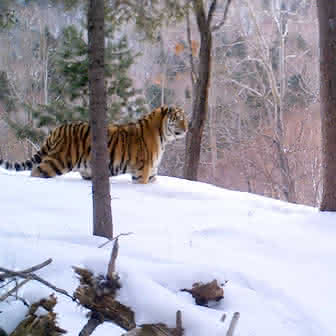}\hfill
        \includegraphics[width=0.16025\linewidth]{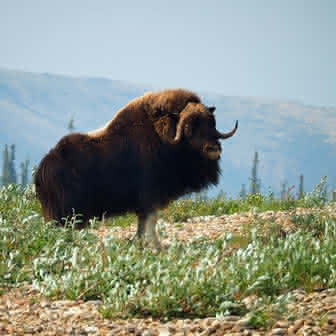}\hfill
        \includegraphics[width=0.16025\linewidth]{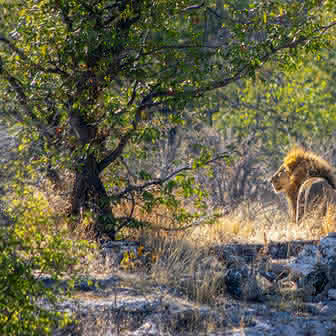}\hfill
        \includegraphics[width=0.16025\linewidth]{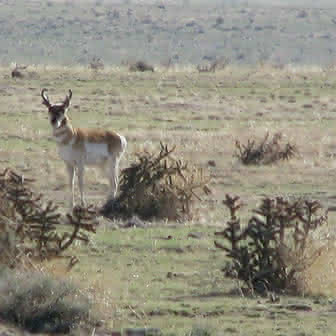}\hfill
        \includegraphics[width=0.16025\linewidth]{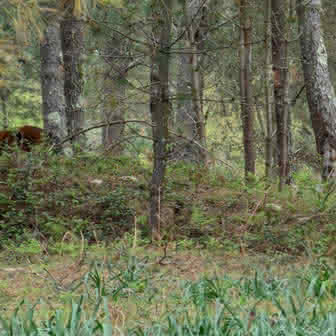}\hfill
        \includegraphics[width=0.16025\linewidth]{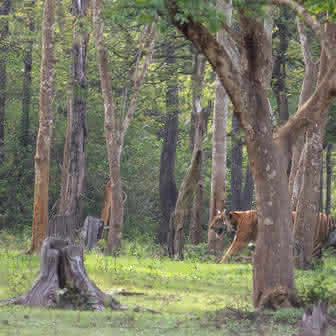}\\
        \vspace{0.1cm}
        \includegraphics[width=0.16025\linewidth]{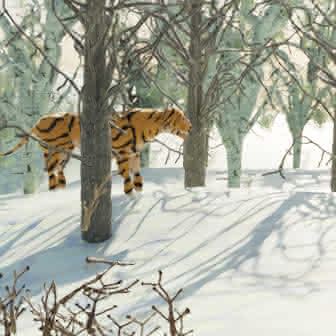}\hfill
        \includegraphics[width=0.16025\linewidth]{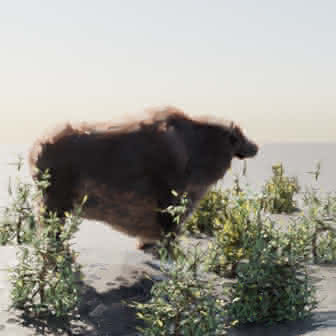}\hfill
        \includegraphics[width=0.16025\linewidth]{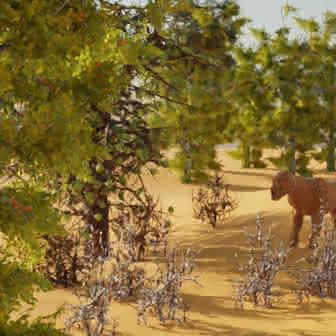}\hfill
        \includegraphics[width=0.16025\linewidth]{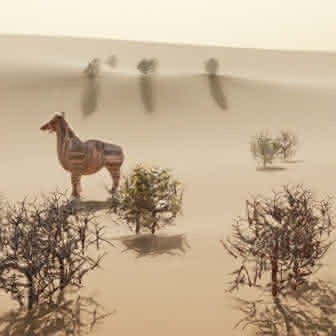}\hfill
        \includegraphics[width=0.16025\linewidth]{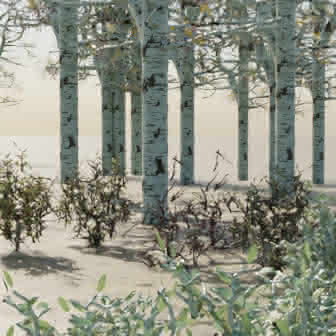}\hfill
        \includegraphics[width=0.16025\linewidth]{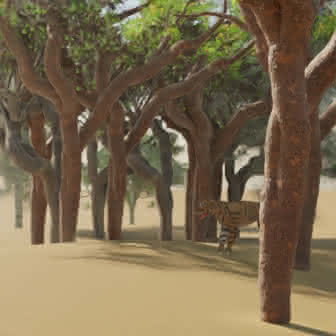}\\
        \vspace{0.1cm}
        \hrule
        \vspace{0.1cm}
        \includegraphics[width=1.0\linewidth]{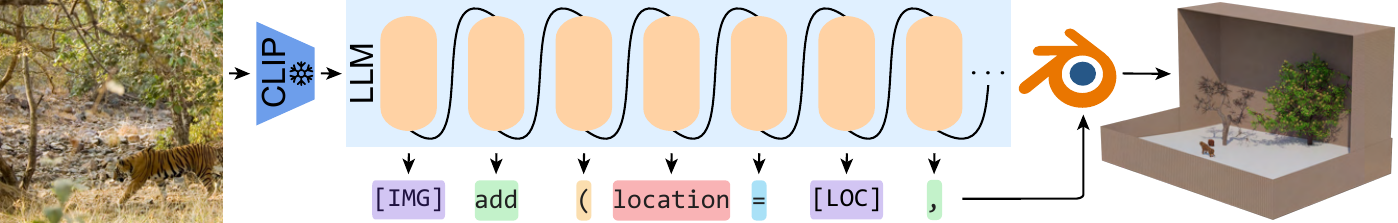}\\
    \end{minipage}
    \captionof{figure}{\teaserCaption}
    \label{fig:teaser}
    \vspace{0.5cm}
}]

\begin{abstract}
The idea of 3D reconstruction as scene understanding is foundational in computer vision.
Reconstructing 3D scenes from 2D visual observations requires strong priors to disambiguate structure.
Much work has been focused on the anthropocentric, which, characterized by smooth surfaces, coherent normals, and regular edges, allows for the integration of strong geometric inductive biases.
Here, we consider a more challenging problem where such assumptions do not hold: the reconstruction of natural scenes containing trees, bushes, boulders, and animals.
While numerous works have attempted to tackle the problem of reconstructing animals \textit{in} the wild, they have focused solely on the animal, neglecting environmental context.
This limits their usefulness for analysis tasks, as animals exist inherently within the 3D world, and information is lost when environmental factors are disregarded.
We propose a method to reconstruct natural scenes from single images.
We base our approach on recent advances leveraging the strong world priors ingrained in Large Language Models and train an autoregressive model to decode a CLIP embedding into a structured compositional scene representation, encompassing both animals and the wild (RAW).
To enable this, we propose a synthetic dataset comprising one million images and thousands of assets.
Our approach, having been trained solely on synthetic data, generalizes to the task of reconstructing animals and their environments in real-world \hbox{images.
We will release our dataset and code to encourage} future research at \url{https://raw.is.tue.mpg.de/}.
\end{abstract}

\section{Introduction}\label{sec:introduction}
The 3D reconstruction of the physical world from visual observations plays a fundamental role in computer vision, providing the foundation for \citet{marr1978representation}'s computational model of visual perception.
The process culminates in a structured 3D representation of the environment.
Compositional 3D reconstructions of scenes, where objects are distinguished into semantic classes, are particularly amenable to analysis, enabling editing and simulation.
When such reconstructions are represented in a compact form that can be used to reproduce the scene, they become expressive models of 3D reality, supporting applications in modeling of physical behavior~\cite{Lake_Ullman_Tenenbaum_Gershman_2017}.

Recent work has built on developments in Large Language Models (LLMs) to reconstruct simple scenes composed of few objects into graphics code~\cite{igllm} or reproduce architectural layouts of indoor scenes, represented in an ad hoc scene language~\cite{avetisyan2024scenescript}.

We consider a more challenging problem, the reconstruction of outdoor natural scenes containing diverse vegetation and animals.
These open settings present unique challenges: unlike man-made scenes, natural environments are harder to interpret, as animals often blend into their surroundings with camouflaging colors and patterns; objects may be positioned at varying distance, some very close and others very far or under a range of lighting conditions; and natural scenes can feature complex interactions between elements, such as trees, animals, and other natural objects.
Unlike \citet{avetisyan2024scenescript} and similar to \citet{igllm}, we reconstruct scenes in graphics code, producing interpretable, editable, and animatable scenes that integrate with existing graphics assets.

While reconstructing natural scenes is itself an unsolved computer-vision problem, we are motivated by the goal of enabling a next-generation computational ethology~\cite{anderson_toward_2014}.
Early vision-aided animal-behavior analysis methods relied on 2D pose observations~\cite{ramanan_animals_video}.
However, 2D pose provides only limited information and, given that the solution is view-dependent, it is typically only applicable in controlled environments for problems like animal-gait analysis from a fixed camera~\cite{KinematicGreyhounds}.
The transition to 3D reconstruction of animals represents a natural progression~\cite{fitzgibbon_dolphin_shape,kanazawa_cats,Zuffi_2017_CVPR}, offering a more comprehensive picture.
However, reconstructing animals in isolation presents limitations for analysis; for example, studying animal behavior in an empty volume cannot account for occlusions, physical boundaries, or natural interactions.
Precise environmental context is useful for understanding animal behavior, yet natural environments pose challenges for both representation and reconstruction.
To date, no work has attempted to concurrently reconstruct both 3D animals and their 3D environment.
Instead, research in recent years has largely focused on creating increasingly detailed 3D representations of isolated animals.
We take a step back, opting instead to work in a complementary direction: rather than pursuing ever-finer animal representations, we prioritize estimating precise layout of the greater scene context with relatively coarse shape representations, to capture the overall  environment.
In this work, we are the first to tackle the challenge of compositionally reconstructing natural scenes from monocular images, presenting the first approach \hbox{to Reconstruct Animals \textit{and} the Wild (RAW); see \cref{fig:teaser}.}

\textbf{Modeling Natural Environments in 3D.}
The 3D reconstruction of natural environments from monocular images presents challenges, stemming both from the fundamental ill-posedness of inverting 2D images into their originating 3D scenes and from a lack of adequate models for representing natural environments.
The natural world is notably more complex and varied than anthropocentric environments with their geometric regularity.
Consequently, modeling natural environments in a manner conducive to analysis is not straightforward.
To address this, we propose a compositional approach that represents environments as ordered sets of objects along with various scene-level attributes.
This object-based representation is interpretable and low-dimensional while abstracting away complexity in a manner that facilitates downstream analysis~\cite{wu2017neural}.

\begin{figure}[t]
\centering

\includegraphics[width=0.16025\linewidth]{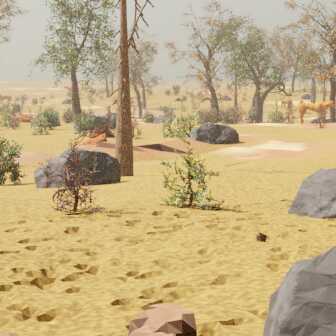}\hfill
\includegraphics[width=0.16025\linewidth]{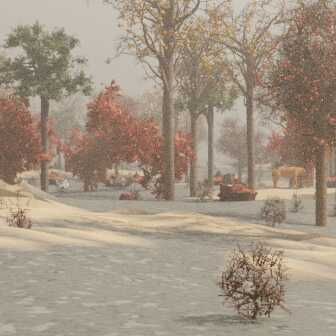}\hfill
\includegraphics[width=0.16025\linewidth]{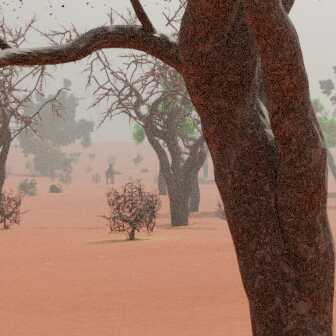}\hfill
\includegraphics[width=0.16025\linewidth]{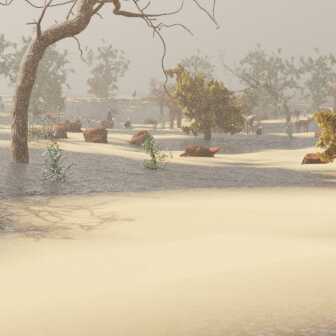}\hfill
\includegraphics[width=0.16025\linewidth]{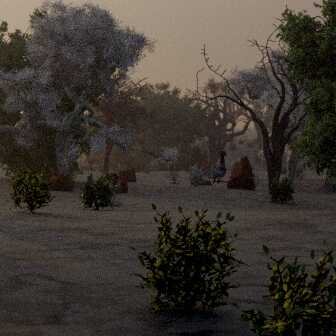}\hfill
\includegraphics[width=0.16025\linewidth]{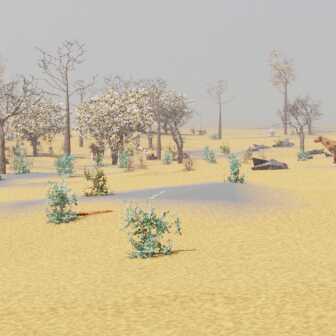}\\
\vspace{0.04cm}
\includegraphics[width=0.16025\linewidth]{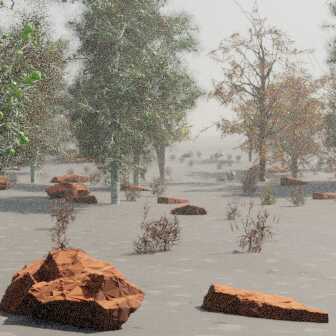}\hfill
\includegraphics[width=0.16025\linewidth]{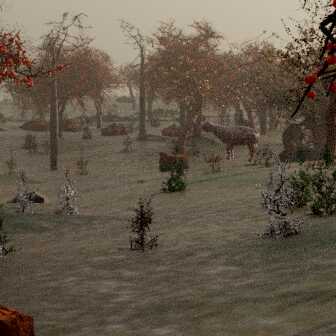}\hfill
\includegraphics[width=0.16025\linewidth]{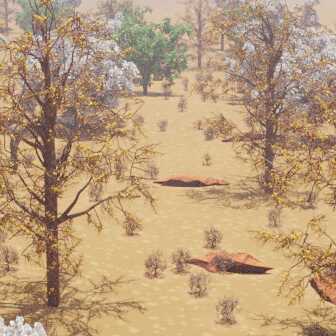}\hfill
\includegraphics[width=0.16025\linewidth]{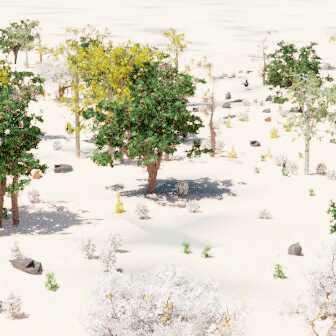}\hfill
\includegraphics[width=0.16025\linewidth]{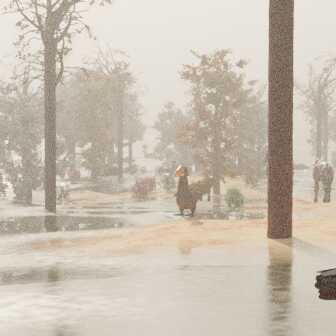}\hfill
\includegraphics[width=0.16025\linewidth]{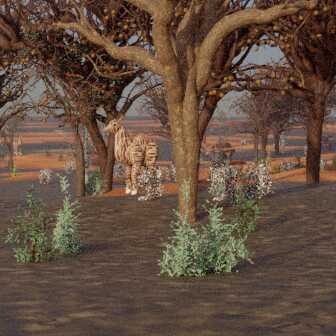}\\
\vspace{0.04cm}
\includegraphics[width=0.16025\linewidth]{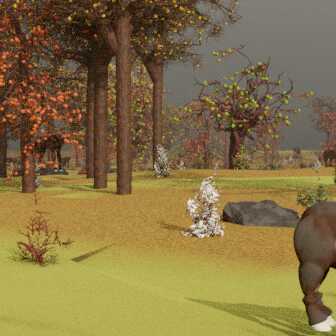}\hfill
\includegraphics[width=0.16025\linewidth]{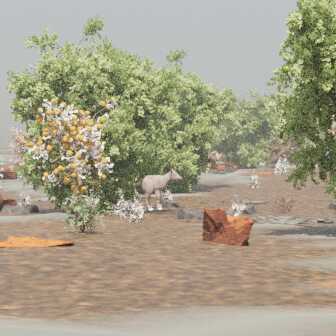}\hfill
\includegraphics[width=0.16025\linewidth]{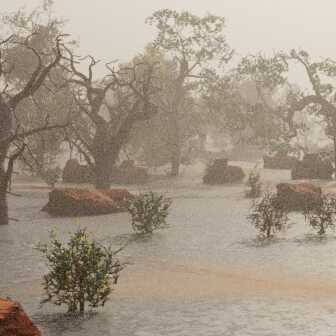}\hfill
\includegraphics[width=0.16025\linewidth]{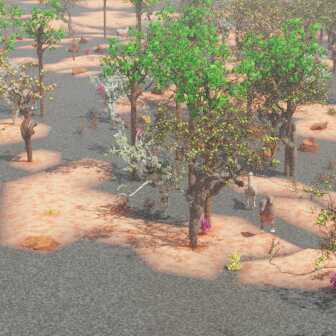}\hfill
\includegraphics[width=0.16025\linewidth]{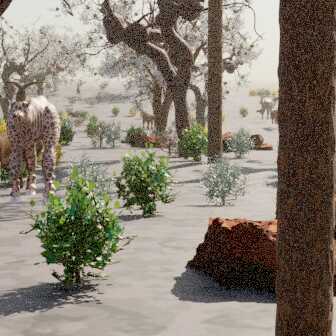}\hfill
\includegraphics[width=0.16025\linewidth]{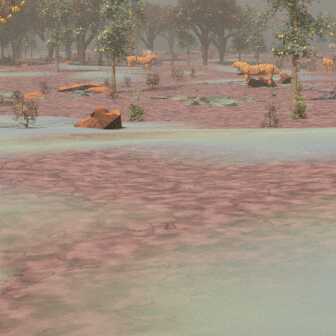}

\caption{\textbf{Dataset Samples.} Training samples from our synthesized dataset. See the Supp.\ Mat.\ for additional visualizations.}
\label{fig:dataset_samples}
\end{figure}

\textbf{Data.}
Teaching a model to decompose single images of animals and their natural environments into structured 3D representations requires broad compositional understanding, yet acquiring suitable training data presents its own challenges.
Because 3D scanning of nature at scale is impractical, here we exploit synthetic-data generation.
Building upon tools introduced by the Infinigen project~\cite{Raistrick_2023_CVPR}, we design a data generator to construct RAW, a million-image dataset comprising both synthetic animals and their environments.
Our scenes encompass a diverse range of elements, including birds, carnivores, herbivores, bushes, boulders, and trees.
See \cref{fig:dataset_samples} for samples.

\textbf{Reconstruction.}
Our goal is to approximately reconstruct 3D animals, scene objects, and layout from a single natural image.
To that end, we design a structured graphics-program representation, or language.
Akin to \cite{igllm}, we train an LLM to decode CLIP~\cite{radford2021learning} image features into graphics code where objects are represented by their asset names.

However, when naively training the language model to produce this sequence, we observe that the model fails to scale to the expanded asset collection---while capturing the layout, it often confuses objects with another (e.g., a tiger with a bush, a bird with a boulder; see \cref{fig:ablation_vis}).
We hypothesize this inconsistency arises due to limitations in train-time supervision.
Built upon a causal LLM, the model operates autoregressively, reconstructing scenes in incremental chunks (\cref{fig:teaser}).
These discrete units, known as tokens, represent bits of text.
During training, language models typically are optimized through a cross-entropy next-token objective, whereby they learn to predict the probabilities of the subsequent token, conditioned on the preceding ones.
Although this discrete text-based representation and supervision excel in capturing distinct categorical attributes such as ``small,'' ``purple,'' ``shiny,'' or ``cube,'' challenges arise when representing naturally continuous quantities~\cite{igllm}.

\begin{figure*}[t]
\centering

\begin{subfigure}[b]{.57\linewidth}
\centering
\raisebox{0.52in}{
\includegraphics[width=\linewidth]{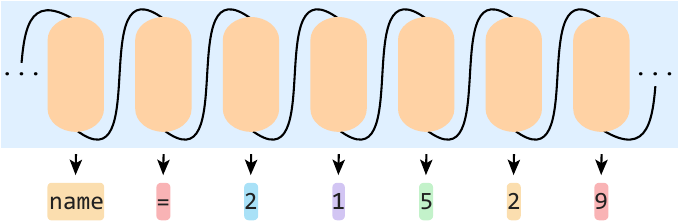}
}
\caption{Discrete Name}
\label{fig:discrete_name}
\end{subfigure}
\hspace{0.5cm}
\begin{subfigure}[b]{.32\linewidth}
\centering
\includegraphics[width=\linewidth]{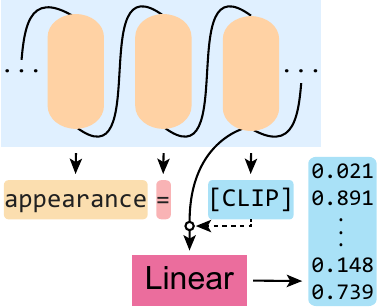}
\caption{Continuous Name}
\label{fig:continuous_name}
\end{subfigure}

\caption{\textbf{CLIP Head.} Rather than teaching the LLM to generate asset names as discrete tokens without a semantically meaningful distance metric, we train the LLM to produce a special token to signal when the LLM hidden state should be projected into a continuous CLIP embedding.}
\label{fig:architecture}
\end{figure*}

Asset names, represented as discrete tokens, lack a meaningful distance metric between one another.
This becomes problematic when estimating identities across a large collection of assets, many with only fine-grained differences in appearance.
We hypothesize that, rather than teaching the LLM to infer exact individual assets by name, the model can be taught to estimate continuous visual appearance, where assets are represented by their CLIP encodings, and prediction is supervised by a loss in semantic CLIP space.
We do so by adding a unique token, \mbox{\texttt{[CLIP]}}, which signals the LLM hidden state should bypass the discretizing tokenization process and, instead, be passed through a linear projection, resulting in a CLIP embedding (\cref{fig:architecture}).

We observe that with the incorporation of the CLIP-projection head, the model demonstrates the ability to scale, estimating objects in scenes featuring much-expanded asset diversity.
Our approach successfully reconstructs animals and their environments in real images.
We will release our dataset and code to encourage further research in this area.

\section{Related Work}\label{sec:related}
\paragraph{Animal Pose and Shape.}
Many works have attempted to estimate animal pose and shape from visual observations, evolving from primitive 2D representations to parametric 3D models.
Early work by \citet{ramanan_animals_video} focused on recovering 2D articulated models of animals from video.
The field later progressed to learning 3D representations, with \citet{fitzgibbon_dolphin_shape} developing a 3D morphable model of a dolphin from images.
\citet{kanazawa_cats} extended this idea, additionally learning to model articulations and pose-dependent deformations.
\citet{Zuffi_2017_CVPR} advanced this further by constructing an articulatable multi-species 3D morphable model from scans of toy animals, used to recover 3D shape and pose of quadrupeds~\cite{Zuffi_2019_ICCV,biggs_2019,biggs2020left,barc2022rueegg}, while others built morphable models to estimate the shape of birds~\cite{badger2020,Wang_2021_CVPR}.

Recent approaches have relatively diverged from a clear progression.
\citet{Kanazawa_2018_ECCV} learned to recover 3D shape and texture of deformable objects from a single image.
\citet{Sanakoyeu_2020_CVPR} adapted 2D dense pose from humans to animals and \citet{Kulkarni_2020_CVPR} developed canonical surface mappings between articulated objects.
\citet{Yang_2022_CVPR} extracted template-free 3D neural models of articulated objects from video, while \citet{jampani_lassie} and \citet{Wu_2023_CVPR} learned articulated 3D shape models using DINO~\cite{Caron_2021_ICCV}-feature-aided part discovery or correspondence.
Sharing an asset-based approach, \citet{liu_casa} estimated 3D animal shape and pose from video by retrieving proximal 3D templates from a collection of video-game assets and deforming the templates to align with extracted features.

\textbf{Inverse-Graphics Approaches.}
The inverse-graphics problem -- the task of \textit{inverting} an image into physical variables that, when rendered, enable reproduction of the observed scene -- has a long history, dating back to Larry Roberts's Blocks-World thesis~\cite{roberts1963machine}.
Considerable efforts have focused on tasks such as estimating object pose~\cite{lepetit2009ep,tejani2014latent,pavlakos20176dof,xiang2017posecnn,wang2021gdr,wang2019normalized,wang2021NeMo,ma2022robust,labbe2020cosypose} and reconstructing shape from single images~\cite{choy20163d,fan2017point,groueix2018papier,mescheder2019occupancy,wang2018pixel2mesh,sitzmann2019scene,park2019deepsdf}. However, works addressing multi-object scenes~\cite{gkioxari2019mesh,denninger20203d,shin20193d} often neglect object semantics and relationships, limiting deeper reasoning.
Holistic 3D-scene understanding aims to reconstruct individual objects along with scene layout.
Initial efforts centered on 3D bounding boxes~\cite{hedau2009recovering,lee2009geometric,mallya2015learning,ren2017coarse,dasgupta2016delay}, with recent advancements emphasizing finer shape reconstruction~\cite{zhang2021holistic,liu2022towards,gkioxari2022learning}.
Relatedly, some methods also involve retrieving CAD or mesh models, followed by 6-DoF pose estimation for objects or scenes~\cite{aubry2014seeing,bansal2016marr,lim2014fpm,tulsiani2015viewpoints,izadinia2017im2cad,huang2018holistic,salas2013slam,kundu20183d,gumeli2022roca,kuo2020mask2cad,kuo2021patch2cad,engelmann2021points}.
In contrast, our work, like \hbox{IG-LLM}~\cite{igllm}, explores the use of LLMs for the inverse-graphics problem, seeking a possibly simpler and more generalizable solution.

\textbf{Learning From Synthetic Data in Vision.}
The use of synthetic data for training transferable vision models has proven highly successful in recent years.
Applications involve learning to detect objects~\cite{Tremblay_2018_CVPR_Workshops}, segment scenes~\cite{Chen_2019_CVPR}, estimate optical flow~\cite{Dosovitskiy_2015_ICCV}, predict depth~\cite{Raistrick_2023_CVPR}, track objects~\cite{Zheng_2023_ICCV}, navigate a robot~\cite{sadeghi_cad2rl,deitke_procthor}, and estimate human pose and shape~\cite{Varol_2017_CVPR,Black_2023_CVPR}.
We further extend this paradigm by learning to extract transferable compositional 3D scene representations of natural scenes using procedurally generated Blender~\cite{blender} scenes, building off tools introduced by the Infinigen project~\cite{Raistrick_2023_CVPR}.

\textbf{LLMs and 3D Understanding.}
Recent applications of LLMs have extended to various 3D-related tasks.
These tasks include 3D question answering~\cite{hong2023_3D_LLM, dwedari2023generating}, planning~\cite{hong2023_3D_LLM,zhang2023building,lv2024gpt4motion}, text-to-3D scene synthesis~\cite{sun2023_3D_GPT,yang2023holodeck,huscenecraft}, procedural model editing~\cite{kodnongbua23reparamCAD}, multi-modal representation learning~\cite{xue2023ulip,hong2023_3D_LLM}, and 3D scene reconstruction from calibrated RGB-D image sequences~\cite{avetisyan2024scenescript}.
These applications demonstrate the wide applicability of LLMs to tasks not traditionally considered text-based.
We continue along the line of IG-LLM~\cite{igllm} and employ an LLM to decode CLIP embeddings into structured 3D-scene representations.

\section{Method}\label{sec:method}
\subsection{Preliminaries}\label{ssec:preliminaries}
\textbf{Autoregressive Language Generation.}\label{sssec:llm}
Causal language models generate text in an autoregressive manner, proceeding chunk by chunk.
Each generated chunk, known as a token, is conditioned on the preceding sequence of generated chunks.
Individual tokens represent bytes or one or more characters~\cite{sennrich-etal-2016-neural}.
The models are typically trained with only a next-token prediction objective~\cite{bengio_plm}, conditioned on the sequence of previously observed tokens:
\begin{equation}\label{eq:next-token}
p(x) = \prod_{i=1}^n p\left(s_i|s_1,\ldots s_{i-1}\right)
\end{equation}
The loss function applied is cross-entropy over the predicted token probabilities.

\textbf{Inverse Graphics With LLMs.}\label{sssec:ig_llm}
LLMs are known for their robust zero-shot generalization capabilities~\cite{Radford2019LanguageMA,Brown2020LanguageMA,Alayrac2022FlamingoAV}, owing to their scale in parameters and the vast amounts of data on which they are trained.
Departing from traditional approaches, the success of LLMs to diverse tasks stems from their training on large and diverse datasets with a simple objective, followed by fine-tuning on smaller, task-specific datasets.
This contrasts with previous paradigms that relied heavily on increasing task-specific data for performance improvements.

Motivated by the remarkable generalization ability of LLMs, IG-LLM~\cite{igllm} treats inverse graphics as LLM-backed inductive program synthesis.
It employs an LLM, aligns a CLIP~\cite{radford2021learning} vision encoder to its token space as a visual tokenizer, and finetunes it on simple demonstrations of images paired with graphics programs, teaching the LLM to decode CLIP embeddings into structured code representations that can be used to reproduce the observed scene in a standard 3D graphics engine.
The demonstrations are produced using procedurally generated images.

\textbf{Continuous-Parameter Estimation in LLMs.}\label{sssec:continuous_parameter}
Tokens are discrete entities, and the cross-entropy loss applied does not impose any particular ordering.
In this loss space, a `4' token is equally distant from a `5' as it is from an `8.'
The discrete nature of tokens makes it difficult to enforce metric supervision.
IG-LLM~\cite{igllm} addresses this challenge by introducing a numeric module for continuous-parameter estimation.
Rather than passing numbers through the text tokenizer, IG-LLM trains the model to produce a special mask token, \mbox{\texttt{[NUM]}}, indicating that the token embedding should bypass the gradient-breaking token discretization and be processed by an MLP to produce a continuous, gradient-preserving parameter estimate.
By circumventing tokenizer discretization, IG-LLM maintains end-to-end differentiability, facilitating the use of metric supervision on produced floats.
This adaptation leads to stronger parameter-space generalization and improved training dynamics.
We adopt a similar approach of using a special token to signal the re-routing of a token embedding.
See \cref{ssec:raw} for further details on our design decisions.

\subsection{Base Architecture}\label{ssec:architecture}
We adopt the framework established by IG-LLM and base our architecture on an instruction-tuned version~\cite{vicuna2023}\footnote{\url{https://huggingface.co/lmsys/vicuna-7b-v1.3}} of LLaMA-7b~\cite{touvron2023llama}, incorporating a frozen CLIP~\cite{radford2021learning} visual tokenizer\footnote{\url{https://huggingface.co/openai/clip-vit-large-patch14-336}} and applying a learnable linear projection to link the vision embeddings with the word-embedding space of the LLM.
Following IG-LLM's coarse vision--language alignment strategy, we pre-train the projector using image--caption pairs sourced from the Conceptual Captions dataset (CC3M)~\cite{sharma2018conceptual}.
See also IG-LLM~\cite{igllm} for additional details on this base setup.

\subsection{Data-Generation Setting}\label{ssec:data_generation}
We design an image--code training-data generator, building upon the tooling of the Infinigen project~\cite{Raistrick_2023_CVPR}.
Infinigen is a procedural data-generation framework designed to create realistic 3D Blender~\cite{blender} scenes of the natural world.
The framework not only generates diverse terrain but produces a broad range of 3D assets to populate these environments, including various types of plants, trees, rocks, and creatures.
These assets are fully parameterized through mathematical rules.
The framework boasts 182 unique procedural asset generators and 1,070 interpretable parameter degrees of freedom, in addition to those parameterized by seeds.

The complexity of the scenes is notable, with the authors reporting an average ``wall time'' of 4.5 hours to create a single image.
Although experiments were conducted on 30k generated images, only ten samples were released, none of which contain 3D-object ground truth.
To effectively model the real world, data generation must be made scalable.

In its public iteration, Infinigen renders an image from a single camera location within each scene; the scene is generated around the camera's field of view to mitigate unnecessary rendering complexity.
This setup presents challenges for randomly placing multiple cameras within the scene ad hoc, as scenes are designed for a single camera perspective.
To improve efficiency, we implement a number of simplifying modifications, including the following:
\begin{itemize}
    \item We limit the generated assets to boulders, bushes, trees, carnivores, herbivores, and birds.
    \item We pre-generate 1,000 instances of each of the above six asset types, totaling 6,000 unique assets.
    \item We generate both high-resolution and low-resolution versions of each asset.
    \item Within each scene, we sample five instances of each tree, bush, boulder, or creature from the pre-generated assets.
    Unlike in Infinigen proper, where assets are individually unique, we instance them.
    \item We populate the entire scene with assets rather than solely in the area surrounding a single camera viewpoint.
\end{itemize}
Following our modifications, we generate 100 images in each of 10,000 distinct scenes, resulting in 1M images.
The scenes are compositional Blender representations.

Only the carnivore, herbivore, and bird assets are natively orientable, that is, have a canonical ``front.''
The trees, boulders, and bushes may have very different visual appearances from different angles, such as a left-leaning tree.
To be able to incorporate and estimate this information, we assign labels to each object based on its yaw (rotation around the vertical axis) relative to the camera.
We divide the yaw into increments of five degrees, resulting in 72 orientations for each object.
This increases the total effective number of assets to 432,000.
In constructing our ground truth, we zero the yaw of the non-orientable objects local to the camera.

\subsection{RAW}\label{ssec:raw}
In this subsection we define our model objective.
We structure the template as follows:
\begin{minted}[breaklines]{python}
set_sun_intensity(0.981)
set_sun_elevation(0.691)
set_sun_size(0.811)
set_camera(88.130)
set_atmospheric_density(0.009)
set_ozone(1.499)
set_sun_rotation(231.110)
set_dust(0.169)
set_sun_strength(0.212)
set_air(0.771)
set_ground([CLIP])
add(pixels=1582, loc=(-0.553, -0.809, -22.591), height=1.365, rotation=[ROT], appearance=[CLIP])
add(pixels=111, loc=(-1.524, -0.939, -30.159), height=1.224, rotation=[ROT], appearance=[CLIP])
\end{minted}
where \mbox{\texttt{[ROT]}} represents a variation of the \mbox{\texttt{[NUM]}} token as applied in IG-LLM, signaling the token embedding to instead be put through an MLP to regress a nine-parameter rotation matrix.
This choice was primarily motivated by reducing code dimensionality to enable the use of up to twenty-five objects per code sequence due to limitations in LLM token context-length allowance, but we also motivate it by the result of an evaluation in IG-LLM demonstrating that employing a continuous representation for rotation estimation enabled greater parameter-space generalization.
The semantic token, \mbox{\texttt{[CLIP]}}, is used to signal whether the token embedding should be projected to CLIP space with a linear layer.
During training, we set the target of the embedding projector to be that of the rendered asset image at the given yaw of the scene asset.
See \cref{fig:architecture} for a visualization.

Scene-level attributes are estimated at the beginning of the sequence prior to objects.
These include sun parameters (intensity, elevation, size, strength, and rotation relative to the camera) and atmospheric conditions (density, ozone, dust content, and air density).
Additionally, a semantic CLIP embedding is estimated to retrieve the ground texture to texture the resulting scene reconstruction.

Objects are ordered in the objective code sequence by the number of pixels they occupy: from the visually largest to the least significant.
In this way, the model is taught to first focus on the most-salient objects before attempting to explain bushes in the background.
See the Supp.\ Mat.\ for a step-wise reconstruction visual, highlighting the ordering estimated by the model.

\subsection{Losses}
In addition to the next-token prediction objective loss applied to the text of the generated code, our use of special tokens for rotation and CLIP-appearance estimation enables and necessitates further supervision.
Following \citet{geist2024learning}, we apply symmetric orthogonalization to our rotation matrices prior to a mean-squared-error loss.
The CLIP-appearance estimation is supervised by a cosine similarity loss between the estimated and target embeddings and an additional regularization term necessary because the similarity loss is vector-norm invariant.
See the Supp.\ Matt.\ for further training details.

\section{Evaluations}\label{sec:evaluations}
\begin{figure}[t]

{
\renewcommand{\arraystretch}{0.4}
\begin{tabular}{@{}c@{\hspace{2pt}}c@{\hspace{1.5pt}}c@{\hspace{1.5pt}}c@{\hspace{1.5pt}}c@{\hspace{1.5pt}}c@{\hspace{1.5pt}}c@{}}
\raisebox{0.4\height}{\rotatebox{90}{\small Input}} & \includegraphics[width=0.1525\linewidth]{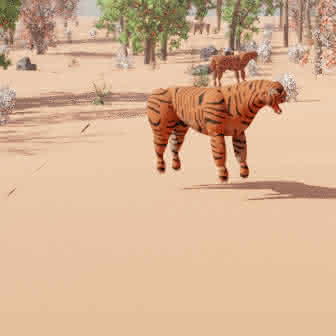} & 
\includegraphics[width=0.1525\linewidth]{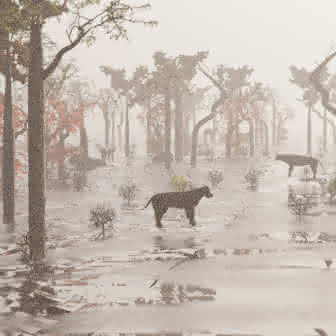} &
\includegraphics[width=0.1525\linewidth]{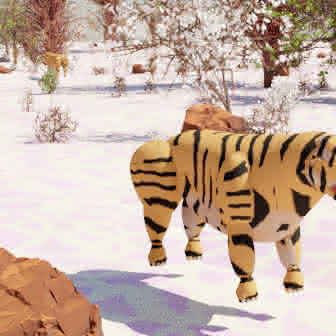} &
\includegraphics[width=0.1525\linewidth]{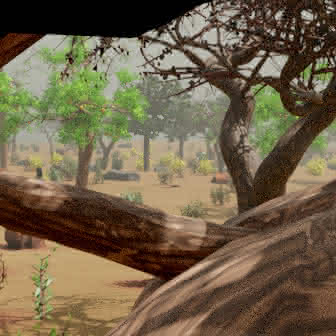} &
\includegraphics[width=0.1525\linewidth]{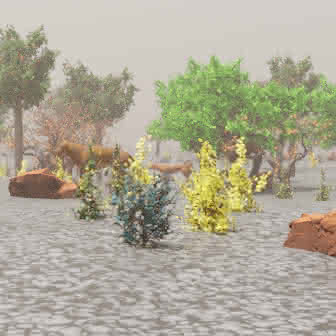} &
\includegraphics[width=0.1525\linewidth]{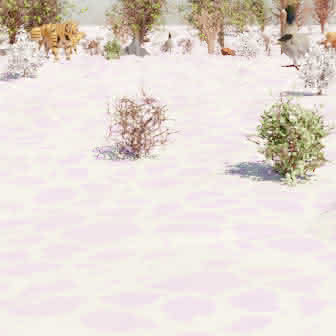} \\
\raisebox{0.075\height}{\rotatebox{90}{\small IG-LLM}} & \includegraphics[width=0.1525\linewidth]{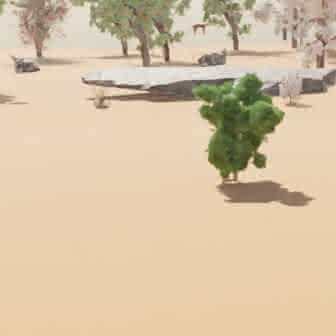} &
\includegraphics[width=0.1525\linewidth]{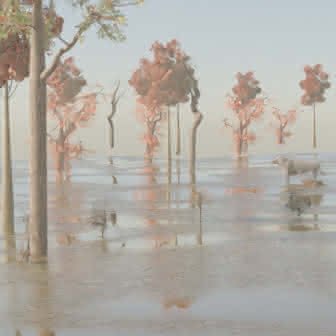} &
\includegraphics[width=0.1525\linewidth]{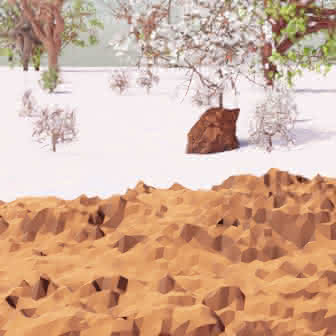} &
\includegraphics[width=0.1525\linewidth]{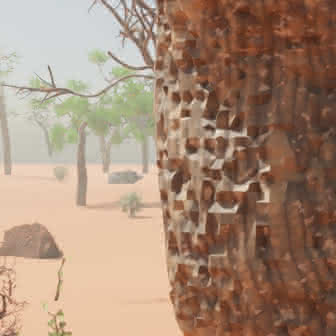} &
\includegraphics[width=0.1525\linewidth]{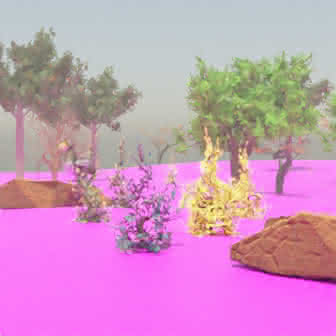} &
\includegraphics[width=0.1525\linewidth]{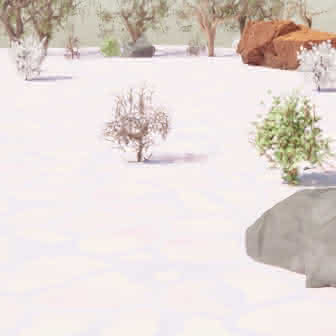} \\
\raisebox{0.25\height}{\rotatebox{90}{\small +CLIP}} & \includegraphics[width=0.1525\linewidth]{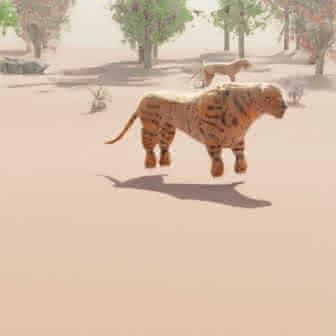} &
\includegraphics[width=0.1525\linewidth]{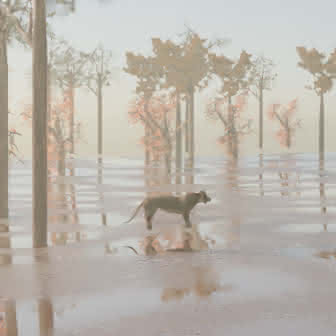} &
\includegraphics[width=0.1525\linewidth]{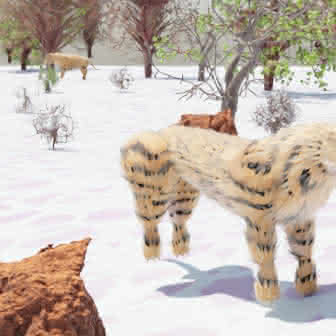} &
\includegraphics[width=0.1525\linewidth]{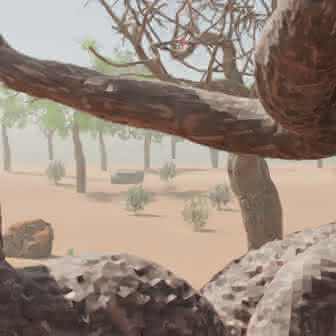} &
\includegraphics[width=0.1525\linewidth]{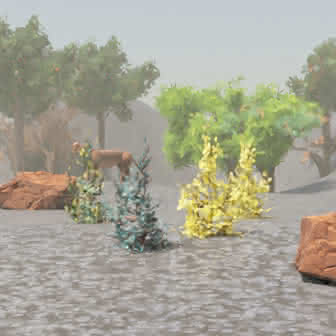} &
\includegraphics[width=0.1525\linewidth]{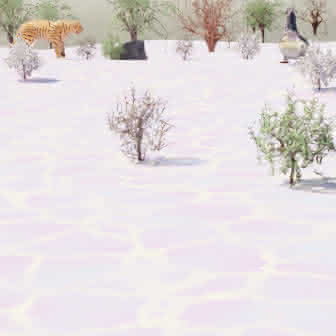}
\end{tabular}
}

\caption{
\textbf{Ablation Visualization.} We observe that, while both the discrete-name IG-LLM baseline and the CLIP-estimation variant well-capture the layout of the in-distribution testing scenes, the discrete variant makes non-interpretable asset-selection errors.
Rather than consistently matching a tiger with another estimated tiger asset, the model confuses it with a bush.
Similarly, a bird is mistaken for a boulder.
Rather than the errors being semantically meaningful misinterpretations, the discrete supervision leads to mistakes that do not make sense.
In contrast, the CLIP-estimation variant consistently identifies objects with aligned interpretations.
}
\label{fig:ablation_vis}
\end{figure}

We evaluate the technical contribution of the proposed method in comparison with the IG-LLM baseline, trained on the RAW dataset.
We begin by measuring the effect of representing assets as semantic CLIP embeddings in \cref{ssec:names}.
In \cref{ssec:fuzz}, we test the effect of conditioning-memorization on generalization ability.
Later, in \cref{ssec:ordering}, we evaluate the effect of introducing additional conditioning into the sequence.
Finally, in \cref{ssec:emb}, we compare alternatives to CLIP for semantic appearance estimation.

\subsection{Metrics}
To evaluate our method quantitatively, we use synthetic data produced using our generator on scenes not observed during training.
We render the estimated scene representation and evaluate it with holistic perceptual metrics against the source image.
Prior to visualization, we warp a ground plane to the estimated object locations using an RBF kernel.
Central to our evaluation is LPIPS~\cite{Zhang_2018_CVPR}, which measures perceptual similarity between two images -- the source and rendered reconstruction -- using features extracted by VGG~\cite{vgg}.
We additionally compute cosine similarity between the input and reconstruction using CLIP~\cite{radford2021learning}, BioCLIP~\cite{stevens2023bioclip}, and DINOv2~\cite{oquab2024dinov2}.
However, we de-emphasize these metrics due to their indirect usage in model ablations but find they offer a complementary perspective to LPIPS. See Supp.\ Mat.\ for object-wise 3D evaluations and a quantitative comparison against YOLOX-6D-Pose~\cite{yolo6d}.
\begin{table}[t]
\centering

{
\setlength{\tabcolsep}{5pt}
\begin{tabular}{lc|ccc}
\toprule
 & $\downarrow$LPIPS & $\uparrow$$S_{\text{CLIP}}$ & $\uparrow$$S_{\text{BioCLIP}}$ & $\uparrow$$S_{\text{DINOv2}}$ \\
\midrule
IG-LLM & 0.720 & 0.748 & 0.421 & 0.833 \\
+ CLIP & 0.654 & 0.806 & \underline{0.537} & \textbf{0.858} \\
+ Fuzz.\ & \underline{0.612} & \underline{0.807} & 0.526 & 0.849 \\
+ Cond.\ & \textbf{0.598} & \textbf{0.815} & \textbf{0.539} & \textbf{0.858} \\
\bottomrule
\end{tabular}
}

\caption{\textbf{Quantitative Ablation Effects}. We observe that each added element contributes to overall model performance. 
$S_{\text{CLIP}}$, $S_{\text{BioCLIP}}$, and $S_{\text{DINOv2}}$ represent respective cosine similarity between embeddings of the source and rendered reconstructions.
}
\label{table:ablation}
\end{table}

\begin{figure*}[t]
\centering

\includegraphics[width=0.0975\linewidth]{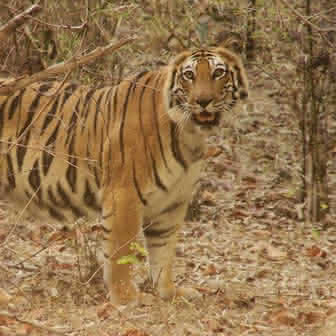}\hfill
\includegraphics[width=0.0975\linewidth]{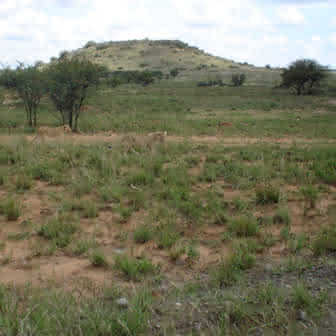}\hfill
\includegraphics[width=0.0975\linewidth]{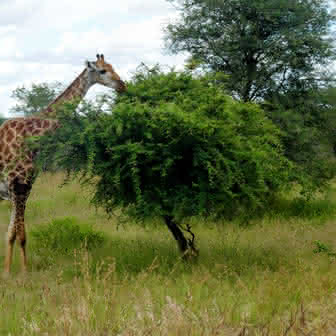}\hfill
\includegraphics[width=0.0975\linewidth]{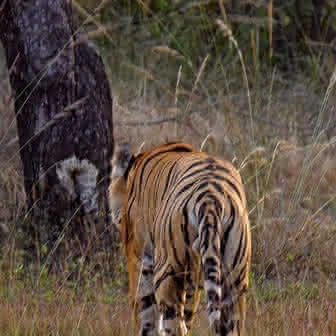}\hfill
\includegraphics[width=0.0975\linewidth]{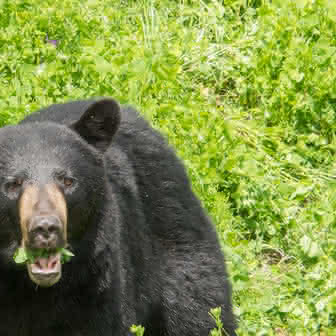}\hfill
\includegraphics[width=0.0975\linewidth]{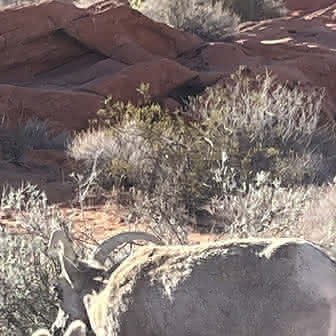}\hfill
\includegraphics[width=0.0975\linewidth]{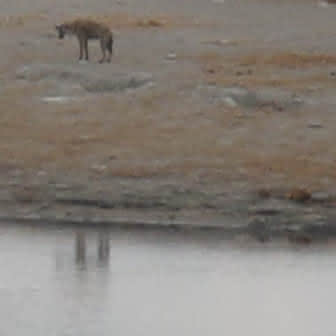}\hfill
\includegraphics[width=0.0975\linewidth]{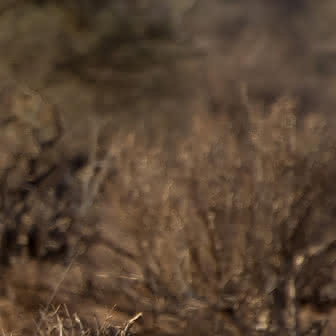}\hfill
\includegraphics[width=0.0975\linewidth]{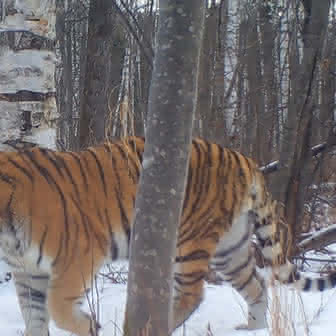}\hfill
\includegraphics[width=0.0975\linewidth]{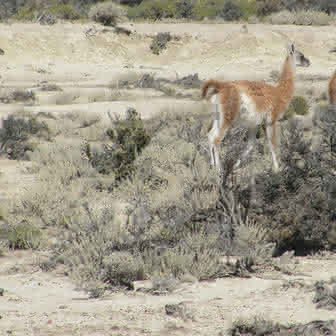}

\includegraphics[width=0.0975\linewidth]{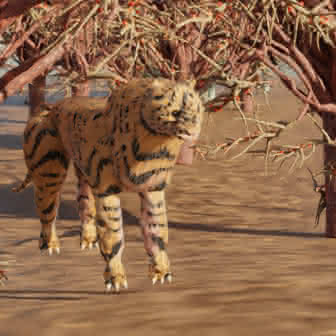}\hfill
\includegraphics[width=0.0975\linewidth]{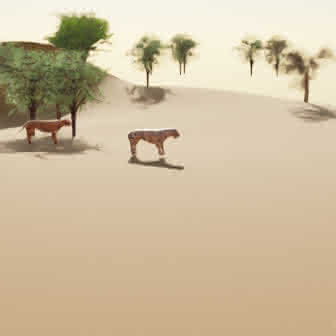}\hfill
\includegraphics[width=0.0975\linewidth]{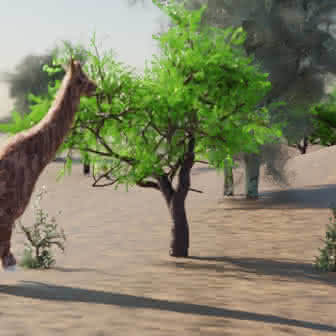}\hfill
\includegraphics[width=0.0975\linewidth]{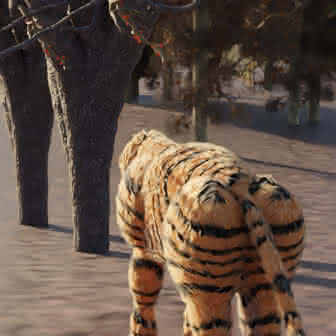}\hfill
\includegraphics[width=0.0975\linewidth]{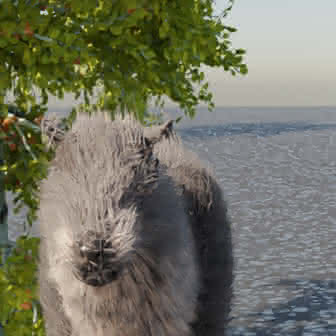}\hfill
\includegraphics[width=0.0975\linewidth]{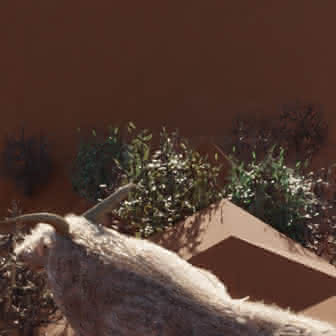}\hfill
\includegraphics[width=0.0975\linewidth]{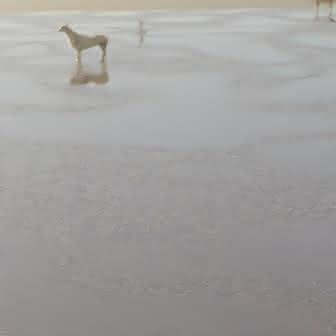}\hfill
\includegraphics[width=0.0975\linewidth]{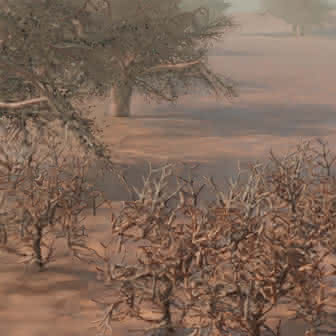}\hfill
\includegraphics[width=0.0975\linewidth]{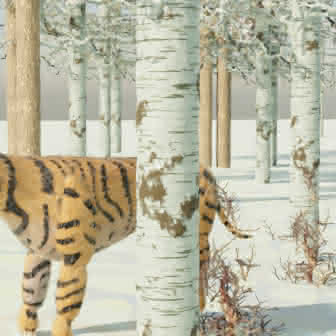}\hfill
\includegraphics[width=0.0975\linewidth]{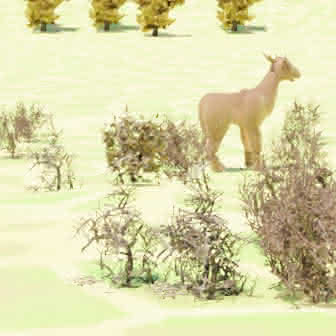}

\includegraphics[width=0.0975\linewidth]{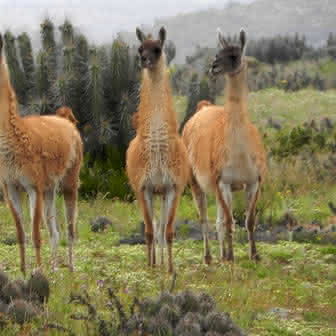}\hfill
\includegraphics[width=0.0975\linewidth]{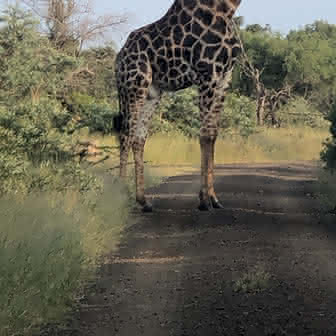}\hfill
\includegraphics[width=0.0975\linewidth]{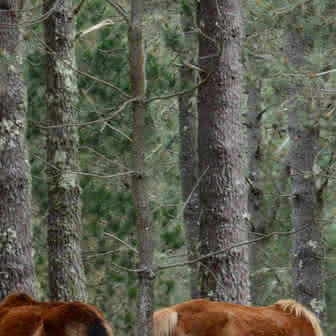}\hfill
\includegraphics[width=0.0975\linewidth]{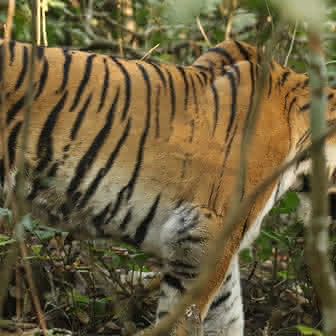}\hfill
\includegraphics[width=0.0975\linewidth]{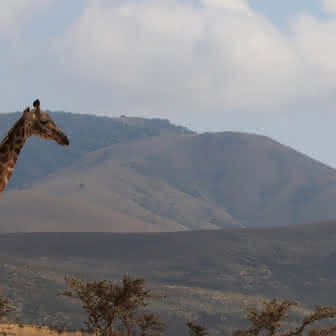}\hfill
\includegraphics[width=0.0975\linewidth]{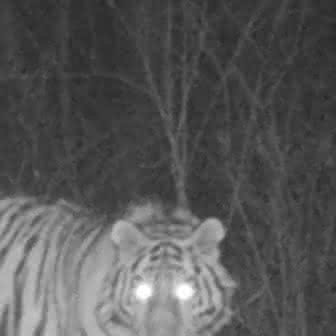}\hfill
\includegraphics[width=0.0975\linewidth]{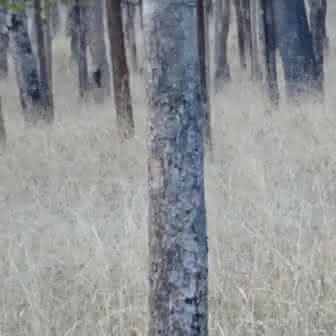}\hfill
\includegraphics[width=0.0975\linewidth]{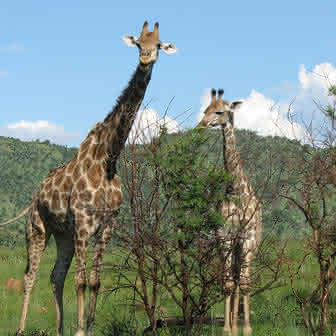}\hfill
\includegraphics[width=0.0975\linewidth]{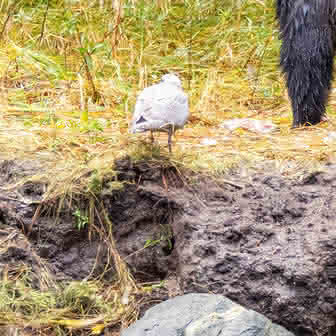}\hfill
\includegraphics[width=0.0975\linewidth]{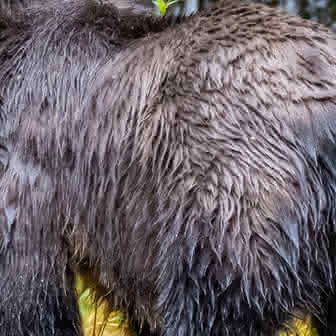}

\includegraphics[width=0.0975\linewidth]{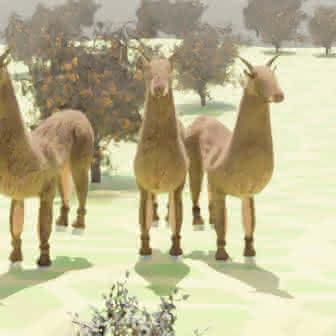}\hfill
\includegraphics[width=0.0975\linewidth]{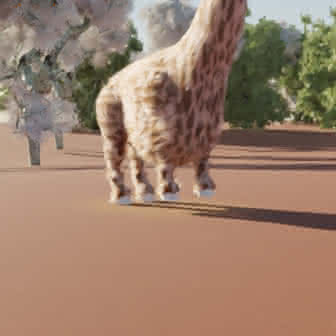}\hfill
\includegraphics[width=0.0975\linewidth]{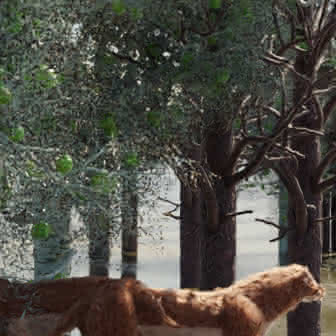}\hfill
\includegraphics[width=0.0975\linewidth]{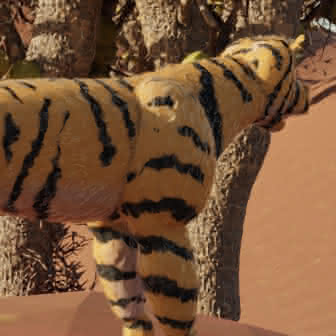}\hfill
\includegraphics[width=0.0975\linewidth]{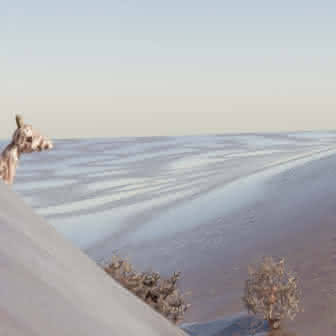}\hfill
\includegraphics[width=0.0975\linewidth]{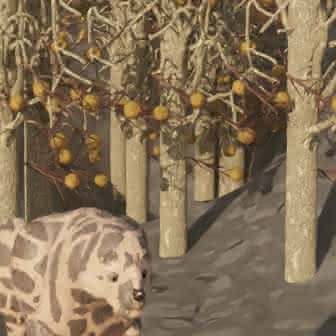}\hfill
\includegraphics[width=0.0975\linewidth]{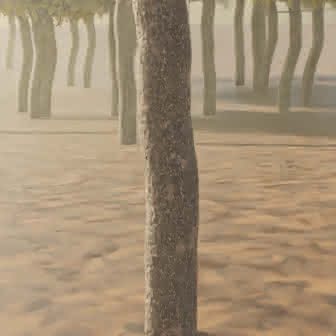}\hfill
\includegraphics[width=0.0975\linewidth]{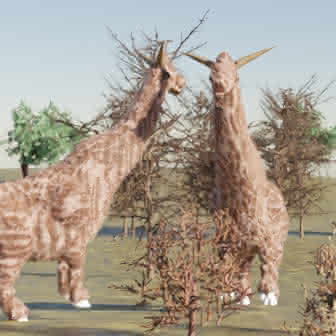}\hfill
\includegraphics[width=0.0975\linewidth]{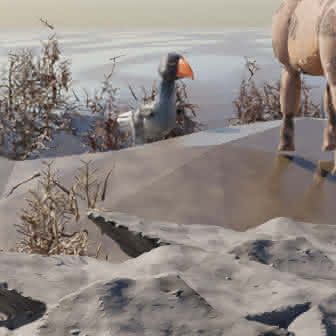}\hfill
\includegraphics[width=0.0975\linewidth]{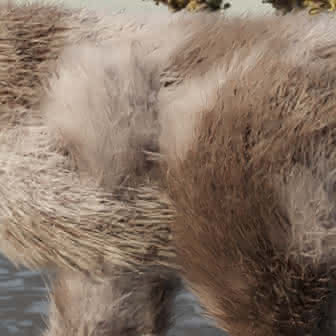}

\caption{\textbf{Additional Reconstructions.} Additional real-world-generalization samples. Note how we can reconstruct scenes where the animal is very far or very close to the camera, with severe occlusion, and in different lighting conditions.}
\label{fig:samples}
\vspace{-0.025cm}
\end{figure*}

\subsection{Discrete Names and Continuous Embeddings}\label{ssec:names}
We begin with evaluating the naive discrete-shape-name version of our pipeline (IG-LLM) trained on our dataset.
In this template, rather than \mbox{\texttt{set\_ground}} and appearance being parametrized by \mbox{\texttt{[CLIP]}} tokens and corresponding embeddings, the assets are named by integers, representing the scene ID in the case of the ground texture.
In this form, asset identities lack continuity and a meaningful metric between them.
See representative examples in \cref{fig:ablation_vis}.

We observe that both models estimate scene layout fairly consistently in-distribution.
However, while many of the assets chosen by both models appear as reasonable approximations, the discrete-name variant has a tendency to confuse assets of similar sizes.
Rather than as another instance of the type of object portrayed, the reconstructions will reference objects of completely different semantic categories: a tiger with a bush or a bird with a boulder.

Transitioning to, instead, estimating semantic asset appearance in the CLIP-estimation variant yields improved results.
Assets are matched semantically, resulting in model explanations with greater perceptual alignment, effectively generalizing.
We find that this change leads to a jump in perceptual alignment across metrics as recorded in \cref{table:ablation}.

\subsection{Value Fuzzing}\label{ssec:fuzz}
We then investigate the effect of memorization of conditioning.
Producing the graphics-code sequence autoregressively, the model conditions the generation of any token on all that precede it.
In the same way, token generations are conditioned on the extracted features of the input image.
In the way that causal language models are trained, the context tokens the model sees are the ground-truth values.

In estimating the value of \texttt{atmospheric\_density}, the model should have all the information necessary to produce the quantity based on only the conditioned-on image features.
However, in practice, if the model knew that the \mbox{\texttt{sun\_intensity}} was 0.981, and that the \mbox{\texttt{sun\_intensity}} is only ever 0.981 in scene 3,389, it might be simpler to memorize a table mapping scene identities to scene attributes.
While some token conditioning is necessary -- the model needs to know that it is estimating \mbox{\texttt{atmospheric\_density}} and hasn't already done so -- conditioning on ground-truth scene-level values may hurt the model's ability to estimate these values in new scenes.
If the training scenes were unique, and each of the million images were from distinct scenes, this could not be expected to be such an issue, but as one hundred images are produced for each scene, we suspect it may harm the model's generalization ability.

We hypothesize that adding a small amount of noise, or ``fuzzing'' to the target scene-level attributes during training will force the model to pay more attention to image features, and learn to better avoid memorization of scenes.
For each value, we add a uniformly distributed $\pm0.5\%$ of noise to each of the scene-level attributes.
The magnitude of the noise added is small enough to not have a noticeable effect on the values themselves but results in a notable improvement in LPIPS (\cref{table:ablation}), supporting our intuition that memorization negatively affects generalization.

\subsection{Additional Conditioning}\label{ssec:ordering}
Next, we evaluate the effect of introducing additional value conditioning to the sequences.
The model, when generating, conditions object predictions off all preceding objects in the sequence.
In the base version, it is able to leverage positional information (\mbox{\texttt{loc}}) and estimated object height (\mbox{\texttt{height}}) to determine where in the sequence it is and what should be produced next.
As the objects are ordered in the objective sequence by the count of their pixels visible in the source image, the model must be able to reason about -- and disentangle -- objects by saliency.
It does not have the option to learn its own ordering, and it must follow along with the GT sequence during training.

We hypothesize that reasoning about this ordering from only the list of what came before is difficult for the model to learn.
Motivated by this, we task the model to additionally estimate the number of pixels visible for each object.
In doing so, it must explicitly model object visibility, and it can also condition off the information during training to reduce uncertainty.
We evaluate adding pixel count to the sequence (\mbox{\texttt{pixels}}), and observe quantitative improvement (\cref{table:ablation}).

\subsection{Choice of Embedding}\label{ssec:emb}
Finally, we explore the use of alternate embeddings for appearance estimation, namely DINOv2~\cite{oquab2024dinov2} and BioCLIP~\cite{stevens2023bioclip}, to determine both their efficacy and the LLM's ability to estimate them effectively.
For clarity, this evaluation does not examine the differences in behavior of alternatives to the CLIP visual tokenizer, which remains fixed throughout the investigation.
DINOv2 is trained without language supervision using a self-supervised learning objective roughly based on masked-image modeling~\cite{He_2022_CVPR}, and BioCLIP is a fine-tuned version of CLIP trained on image–label data of taxonomic species
from the iNaturalist~\cite{Horn_2018_CVPR} and BIOSCAN-1M~\cite{bioscan} databases.

We quantitatively evaluate the variants in terms of the earlier evaluation setting.
Results can be seen in \cref{table:embedding}.
Contrary to our initial speculation, we find that BioCLIP performs worst as the target embedding across metrics.
We suspect that the finetuning applied in training the model hinders its generalization ability.
On the other hand, DINOv2, which we had originally expected to be less natural to the model and more difficult to learn, performed best across metrics.
While performing well in-distribution, we observe that the DINOv2-based model does not effectively generalize to real-world images.
We suggest that, while the features may increase the separability of the assets, the space is not as interpretable to the LLM and it does not learn a general transform.
See the Supp.\ Mat.\ for a perceptual evaluation conducted on real-world images.

\begin{table}[t]
\centering

{
\setlength{\tabcolsep}{5pt}
\begin{tabular}{lc|ccc}
\toprule
 & $\downarrow$LPIPS & $\uparrow$$S_{\text{CLIP}}$ & $\uparrow$$S_{\text{BioCLIP}}$ & $\uparrow$$S_{\text{DINOv2}}$ \\
\midrule
CLIP & \underline{0.598} & \underline{0.815} & \underline{0.539} & \underline{0.858} \\
BioCLIP & 0.676 & 0.795 & 0.512 & 0.833 \\
DINOv2 & \textbf{0.597} & \textbf{0.850} & \textbf{0.603} & \textbf{0.865} \\
\bottomrule
\end{tabular}
}

\caption{\textbf{Embedding Choice.} On our in-distribution quantitative evaluation set, we observe that employing DINOv2 features as the target embedding leads to the strongest performance, while BioCLIP features fare the worst.
}

\label{table:embedding}
\end{table}

\section{Limitations and Future Work}\label{sec:limitations}

While assets produced using Infinigen tooling are parametricaly defined, reconstructing them solely as code presents challenges, as they are frequently the result of complex non-invertible physical processes, including seeded noise operations.
Consequently, while the assets may have compact low-dimensional representations, the parameters are not smooth: small seed changes might result in an object with dramatically different shape or appearance.
Future work could explore a compromise between retrieval and asset generation (predicting some parameters while retrieving others), rely on differentiable proxies for the non-invertible steps, or employ models with fully interpretable parameters.

The current creature-articulation system in Infinigen is non-functional (and, as pictured throughout \citet{Raistrick_2023_CVPR}, all creatures are of static pose, many with feet visibly off the ground).
This restricts the expressivity of the data-generation framework.
Future work may involve inferring full, articulated object pose.

\begin{figure}
\centering

\includegraphics[width=0.16025\linewidth]{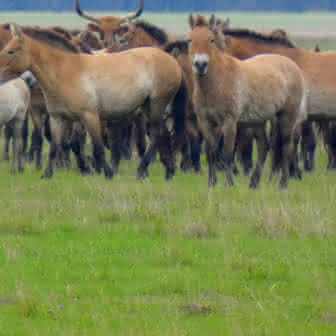}\hfill
\includegraphics[width=0.16025\linewidth]{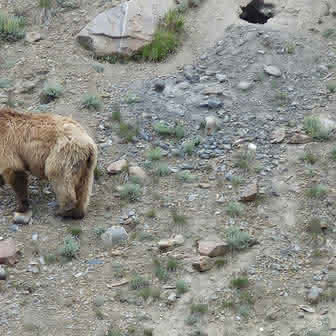}\hfill
\includegraphics[width=0.16025\linewidth]{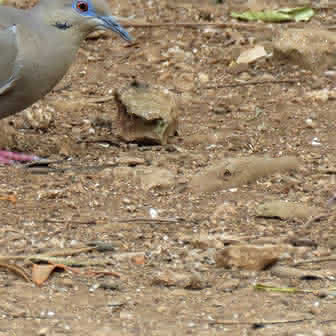}\hfill
\includegraphics[width=0.16025\linewidth]{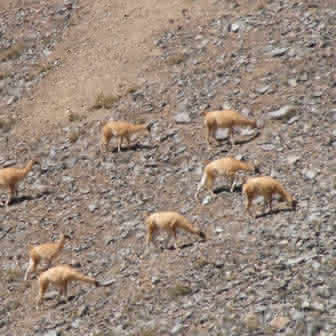}\hfill
\includegraphics[width=0.16025\linewidth]{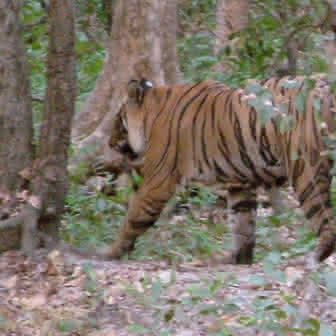}\hfill
\includegraphics[width=0.16025\linewidth]{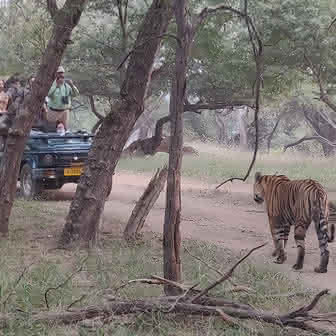}\\
\vspace{0.0375cm}
\includegraphics[width=0.16025\linewidth]{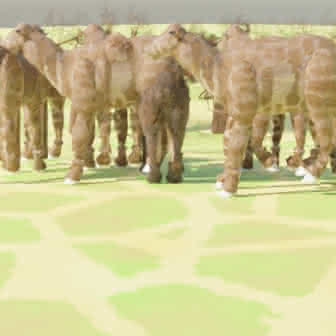}\hfill
\includegraphics[width=0.16025\linewidth]{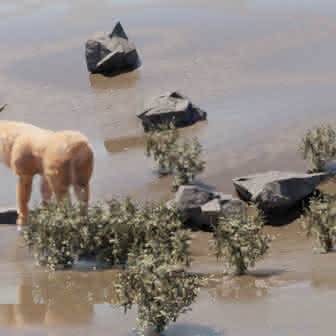}\hfill
\includegraphics[width=0.16025\linewidth]{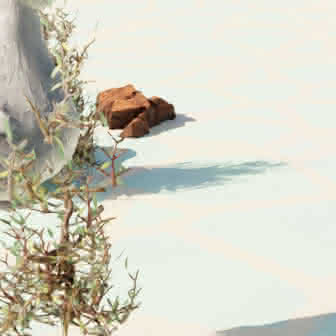}\hfill
\includegraphics[width=0.16025\linewidth]{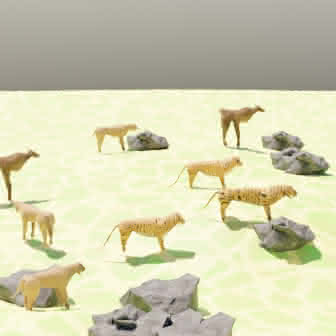}\hfill
\includegraphics[width=0.16025\linewidth]{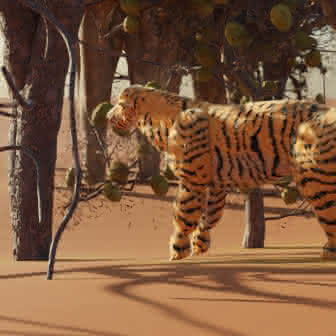}\hfill
\includegraphics[width=0.16025\linewidth]{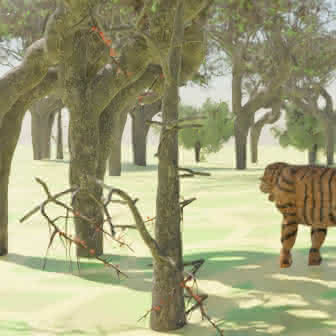}

\caption{\textbf{Limitations Samples.} Our model can struggle reconstructing scenes from images with highly out-of-distribution layout or camera pose.
}
\label{fig:failure}
\end{figure}

While we observe fairly consistent semantically aligned reconstruction of real-world environments, our model can struggle to reconstruct images of scenes with highly out-of-distribution configurations, such as those in which the pose of the camera is outside the distribution seen during training.
We suspect that some such generalization issues might be abated with better layout sampling during data-generation, but without a more-diverse pool of assets, the model will not be able to sufficiently explain all aspects of the scene, such as a vehicle on the road (\cref{fig:failure}).

\section{Conclusion}\label{sec:conclusion}
Our investigation represents the first compositional reconstruction of natural scenes that captures both animals and their natural environments, bridging the gap between reconstructing animals \textit{and} the wild.

In summary, we make the following key contributions:

First, we identify and address a fundamental limitation in scaling scene reconstruction to expansive asset collections. By teaching an LLM to ``name'' objects in terms of semantic CLIP appearance rather than as discrete tokens, we overcome limitations inherent in pure token-based supervision.

Second, we introduce a million-image synthetic dataset built on tooling introduced in the Infinigen project~\cite{Raistrick_2023_CVPR}, addressing data limitations.
Despite training exclusively on this synthetic data, our approach successfully generalizes to reconstructing natural scenes from single images.

Finally, by enabling comprehensive scene reconstruction that includes both animals and their environmental context, we lay the groundwork for a next generation of computational ethology. This opens new possibilities for automated interpretation of animal behavior grounded in their habitat.

\section{Acknowledgements}\label{sec:acknowledgements}
We thank Tomasz Niewiadomski, Taylor McConnell, and Tsvetelina Alexiadis for study assistance, Nikos Athanasiou for discussions, and Rick Akkerman for proofreading.\\

\noindent\textbf{Disclosure.}
MJB has received research gift funds from Adobe, Intel, Nvidia, Meta/Facebook, and Amazon.  MJB has financial interests in Amazon and Meshcapade GmbH.  While MJB is a co-founder and Chief Scientist at Meshcapade, his research in this project was performed solely at, and funded solely by, the Max Planck Society. SZ is supported by PNRR FAIR Future AI Research (PE00000013), Spoke 8 Pervasive AI (CUP H97G22000210007) and NBFC National Biodiversity Future Center (CN00000033), Spoke 4 (CUP B83C22002930006) under the NRRP MUR program by NextGenerationEU.

\clearpage

{
    \small
    \bibliographystyle{ieeenat_fullname}
    \bibliography{main}
}

\end{document}